\pdfoutput=1
\documentclass[11pt]{article}

\usepackage[preprint]{acl}
\usepackage{booktabs}
\usepackage{times}
\usepackage{latexsym}

\usepackage{multirow}
\usepackage{amsmath}
\usepackage{amsfonts, amssymb}
\usepackage{bbding}
\usepackage{amsmath} % for enhanced math environments
\usepackage{breqn} % for automatic line breaking
\usepackage{mdframed}
\usepackage[T1]{fontenc}
\usepackage[utf8]{inputenc}

\usepackage{microtype}

\usepackage{inconsolata}
\usepackage{graphicx}
\usepackage{color}

\newcommand\blfootnote[1]{% 
    \begingroup 
    \renewcommand\thefootnote{}\footnote{#1}% 
    \addtocounter{footnote}{-1}% 
    \endgroup 
}

\title{MatchTime: Towards Automatic Soccer Game Commentary Generation}

\author{Jiayuan Rao$^{*}$, Haoning Wu$^{*}$, Chang Liu, Yanfeng Wang$^{\dagger}$, Weidi Xie$^{\dagger}$\\
  School of Artificial Intelligence, Shanghai Jiao Tong University, China\\
  \texttt{\{jy\_rao, whn15698781666, liuchang666, wangyanfeng622, weidi\}@sjtu.edu.cn} \\
  \url{https://haoningwu3639.github.io/MatchTime/}
}

\begin{document}
\maketitle

 \blfootnote{
 \hspace{-0.671cm} *: These authors contribute equally to this work. 
 \\
 $\dagger$: corresponding author.
}

\begin{abstract}
Soccer is a globally popular sport with a vast audience, 
in this paper, we consider constructing an automatic soccer game commentary model to improve the audiences' viewing experience.
In general, we make the following contributions:
{\em First}, observing the prevalent video-text misalignment in existing datasets, we manually annotate timestamps for 49 matches, establishing a more robust benchmark for soccer game commentary generation, termed as {\em SN-Caption-test-align};
{\em Second}, we propose a multi-modal temporal alignment pipeline to automatically correct and filter the existing dataset at scale, creating a higher-quality soccer game commentary dataset for training, denoted as {\em MatchTime};
{\em Third}, based on our curated dataset, we train an automatic commentary generation model, named \textbf{MatchVoice}. 
Extensive experiments and ablation studies have demonstrated the effectiveness of our alignment pipeline, and training model on the curated dataset achieves state-of-the-art performance for commentary generation, showcasing that better alignment can lead to significant performance improvements in downstream tasks.

 % and leveraging existing multimodal generative models
% To address this issue and promote the development and popularization of soccer, we aim to build a higher-quality dataset to train a stronger soccer commentary model.
% Concretely, we discover the common visual-language misalignment problem in existing soccer game datasets and manually correct 49 matches to serve as a more precise benchmark.
% Subsequently, we make full use of video, audio and text commentary from soccer matches to perform automatic temporal video-text alignment in a coarse-to-fine manner.
% Inspired by existing video understanding models, we further train a powerful LLaMA-style commentary model based on this higher-quality dataset.
% Extensive experiments and ablation studies have demonstrated that better alignment leads to significant performance improvements in downstream tasks, showcasing that our model achieves state-of-the-art commentary performance.

% \haoning{Need a major overhaul}

\end{abstract}
\begin{figure*}[t]
   \centering
   \includegraphics[width=\textwidth]{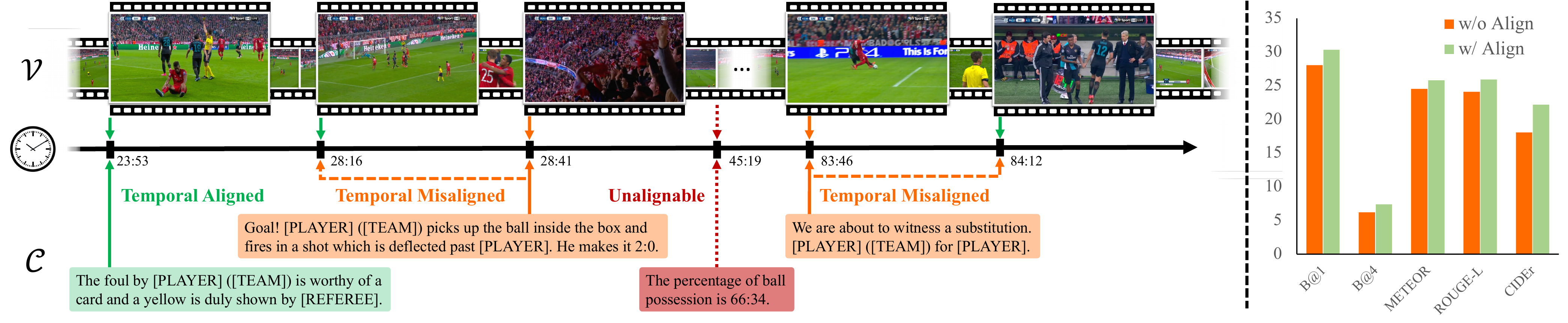}
\caption{\textbf{Overview.}
   (a) {\em Left}: Existing soccer game commentary datasets contain significant misalignment between visual content and textual commentaries.
   We aim to align them to curate a better soccer game commentary benchmark.
   (b) {\em Right}: While evaluating on manually aligned videos, 
   existing models can achieve better commentary quality in a zero-shot manner.
   (The temporal window size is set to 10 seconds here.)
  }
 \vspace{-4pt}
  \label{fig:teaser}
\end{figure*}

\section{Introduction}
Soccer, as one of the most popular sports globally, has captivated over 
5 billion~\cite{FIFA}
% 5 billion~\cite{FIFA}
viewers with its dynamic gameplay and intense moments. 
Commentary plays a crucial role in improving the viewing experience, providing context, analysis, and emotional excitement to the audience. 
However, creating engaging and insightful commentary requires significant expertise and can be resource-intensive. 
In recent years, advancements in artificial intelligence, particularly in foundational visual-language models, have opened new possibilities for automating various aspects of content creation. 
This paper aims to develop an high-quality, automatic soccer commentary system. 

In the literature on video understanding, there has been relatively little attention on sports videos. 
Pioneering work such as SoccerNet~\cite{giancola2018soccernet1} introduces the first soccer game dataset, containing videos of 500 soccer matches. 
Subsequently, SoccerNet-Caption~\cite{densecap} compiles textual commentary data for 471 of these matches from the Internet, establishing the first dataset and benchmark for soccer game commentary. 
However, upon careful examination, we observe that the quality of existing data is often unsatisfactory. 
For instance, as illustrated in Figure~\ref{fig:teaser} (left), since the textual commentaries are often collected from the text live broadcast website, there can be a delay with respect to the visual content, leading to prevalent misalignment between textual commentaries and video clips.

In this paper, we start by probing the effect of the above-mentioned misalignment on the soccer game commentary systems. 
Specifically, we manually correct the timestamps of commentaries for 49 matches in the SoccerNet-Caption test set to obtain a new benchmark, termed as {\em SN-Caption-test-align}. 
With manual check, we observe that these misalignments can result in temporal offsets for up to \textbf{152} seconds, with an average absolute offset of \textbf{16.63} seconds. 
As depicted in Figure~\ref{fig:teaser}~(right), after manual correction, pre-trained off-the-shelf SN-Caption model~\cite{densecap} has exhibited large performance improvements, underscoring the effect of temporal alignment. 

To address the aforementioned misalignment issue between textual commentaries and visual content, we propose a two-stage pipeline to automatically correct and filter the existing commentary training set at scale. 
We first adopt WhisperX~\cite{bain2022whisperx} to extract narration texts with corresponding timestamps from the background audio, which are then summarised into event descriptions by LLaMA-3~\cite{llama3} at fixed intervals. 
Subsequently, we utilize LLaMA-3 to select the most appropriate time intervals based on the similarity between these timestamped event descriptions and textual commentaries. 
Given such an operation only provides rough alignment, we further align the video and commentary by training a multi-modal temporal alignment model on a small set of manually annotated videos. 

Our alignment pipeline enables to significantly mitigate the temporal offsets between the visual content and textual commentaries, resulting in an higher-quality soccer game commentary dataset, named \textbf{MatchTime}. With such a curated dataset, we further develop a video-language model by connecting visual encoders with language model, termed as \textbf{MatchVoice}, that enables to generate accurate and professional commentaries for soccer match videos. 
Experimentally, we have thoroughly investigated the different visual encoders, demonstrating state-of-the-art performance in both precision and contextual relevance.

To summarize, we make the following contributions:
(i) we show the effect of misalignment in automatic commentary generation evaluation by manually correcting the alignment errors in 49 soccer matches,
which can later be used as a new benchmark for the community, 
termed as {\bf SN-Caption-test-align}, as will be detailed in Sec.~\ref{sec:benchmark_curation}; 
(ii) we further propose a multi-modal temporal video-text alignment pipeline that corrects and filters existing soccer game commentary datasets at scale, resulting in an high-quality training dataset for commentary generation, named {\bf MatchTime}, as will be detailed in Sec.~\ref{sec:aligning_commentary_and_videos};
(iii) we present a soccer game commentary model named \textbf{MatchVoice}, establishing a new state-of-the-art performance for automatic soccer game commentary generation, as will be detailed in Sec.~\ref{sec:automatic_soccer_game_commentary}.

\begin{figure}[t]
  \centering
  \includegraphics[width=0.48\textwidth]{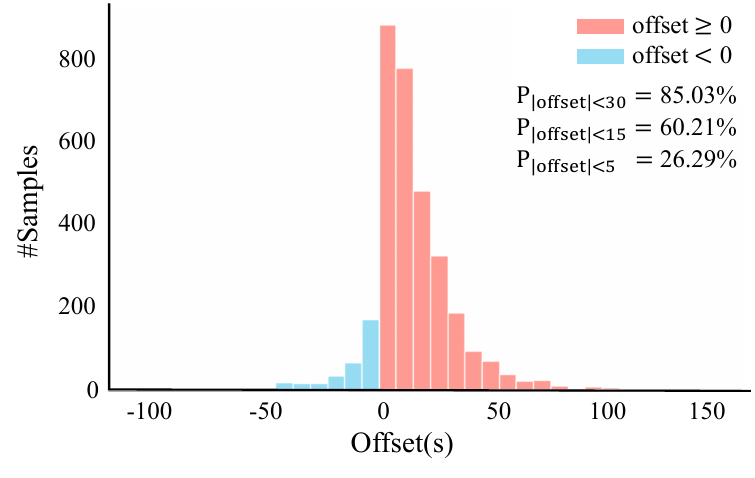} \\
  \vspace{-10pt}
  \caption{\textbf{Distribution of temporal offsets} in our manually corrected {\bf SN-Caption-test-align}.
  Through manual annotation, we find that the temporal discrepancy between the textual commentary and the visual content in the existing benchmark can even exceed 100 seconds. }
  \vspace{-10pt}
 \label{fig:data_preparation}
\end{figure}

\begin{figure*}[t]
  \centering
  \includegraphics[width=\textwidth]{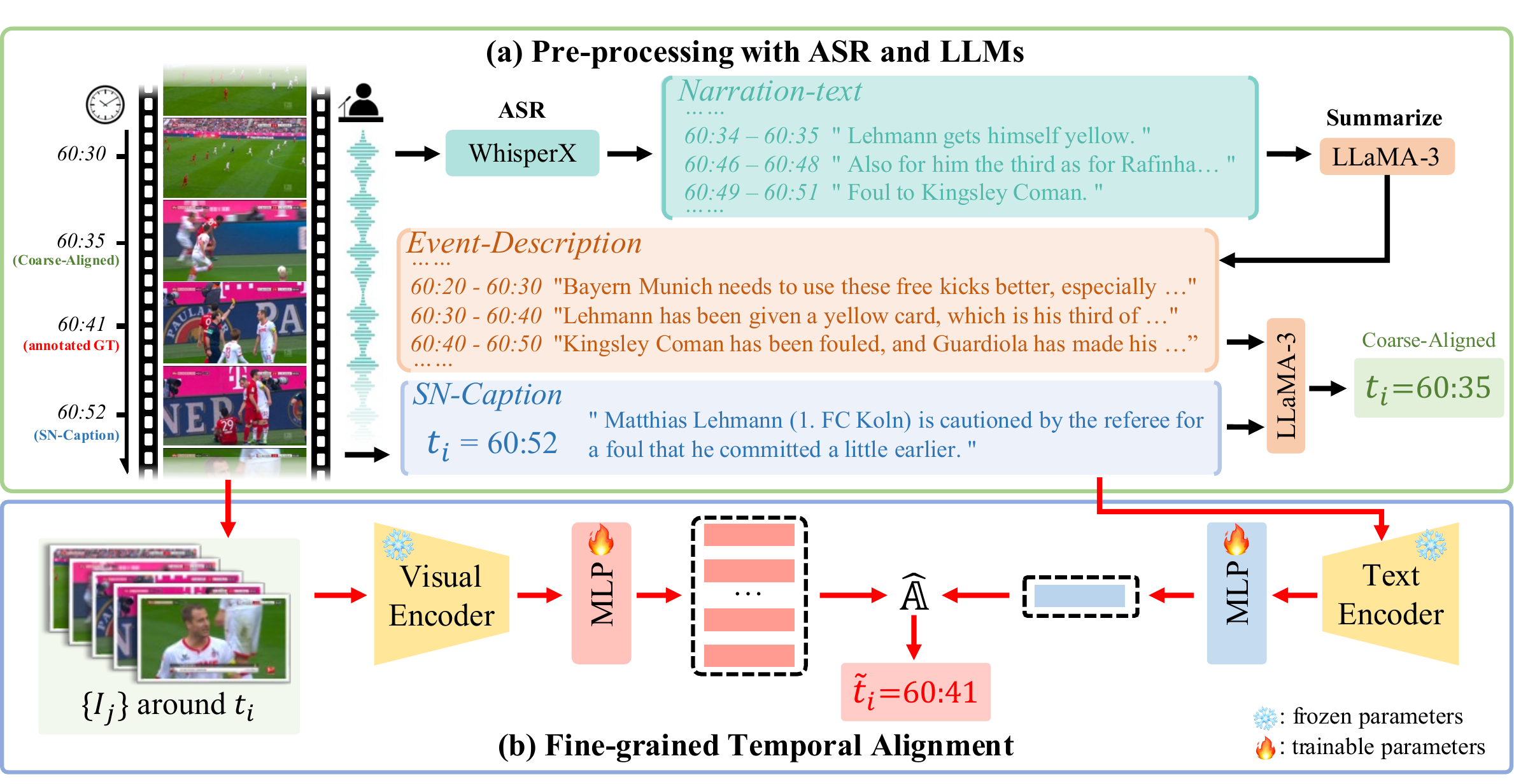} \\
  \vspace{-4pt}
  \caption{\textbf{Temporal Alignment Pipeline}. 
  (a) Pre-processing with ASR and LLMs: We use WhisperX to extract narration texts and corresponding timestamps from the audio, and leverage LLaMA-3 to summarize these into a series of timestamped events, for data pre-processing.
  (b) Fine-grained Temporal Alignment: We additionally train a multi-modal temporal alignment model on manually aligned data, which further aligns textual commentaries to their best-matching video frames at a fine-grained level.
  % \weidi{to check the caption, there is no training-free coarse-grained alignment any more.}
  }
 \vspace{-0.2cm}
 \label{fig:alignment}
\end{figure*}

\section{Benchmark Curation}
\label{sec:benchmark_curation}
To probe the effect of misalignment on the performance of soccer game commentary models, we have manually annotated the timestamps of textual commentaries for 49 matches in the test set of SoccerNet-Caption, resulting in a new benchmark, denoted as {\bf SN-Caption-test-align}.

\vspace{2pt} \noindent{\textbf{Mannual Annotations.}}
We recruit 20 football fans to manually align textual commentaries with video content for 49 matches from the test set of SoccerNet-Caption~\cite{densecap}, following several rules:
(i) Volunteers should watch the entire video, and adjust the timestamps of original textual commentaries to match the moments when events occur;
(ii) To ensure the continuity of actions such as {\em shots}, {\em passes}, and {\em fouls}, the manually annotated timestamps are adjusted 1 second earlier to capture the full context;
(iii) For scenes with replays, the timestamp of the event's first occurrence is marked as the corresponding commentary timestamp to maintain visual integrity and consistency. 

Here, our annotated dataset serves two purposes:
{\em first}, it acts as a more accurate benchmark for evaluating soccer game commentary generation; {\em second}, it can be used as supervised data for training and evaluating temporal alignment pipelines.

\vspace{2pt} 
\noindent{\textbf{Data Statistics.}}
After manually annotating the test set videos, 
we obtain a total of 3,267 video-text pairs. 
As depicted in Figure~\ref{fig:data_preparation}, 
we show the temporal offset between the original noisy timestamps of the textual commentary and the manually annotated ground truth, which ranges from -108 to 152 seconds, with an average offset of 13.85 seconds and a mean absolute offset of 16.63 seconds. 
Only 26.29\%, 60.21\%, 74.96\%, and 85.03\% of the data falls within 10s, 30s, 45s, and 60s windows around the key frames, respectively. 
This highlights the severe misalignment in existing datasets, which will potentially confuse the model training for automatic commentary generation.
\section{Aligning Commentary and Videos}
\label{sec:aligning_commentary_and_videos}
In this section, we develop an automatic pipeline for aligning the timestamps of given textual commentaries to the corresponding video content in existing soccer game commentary datasets.
In Sec.~\ref{sec:temporal_alignment}, we start with the problem formulation for temporal alignment, and subsequently, in Sec.~\ref{sec:coarse_to_fine_temporal_alignment}, we elaborate on the details of our proposed multi-modal temporal alignment pipeline.

% In this section, we first describe the process of manual annotations and demonstrate the necessity of temporal alignment.
% Subsequently, we present a comprehensive statistical analysis of the aligned dataset, namely {\bf MatchTime}.

%Considering the temporal misalignment in existing soccer game commentary datasets, 
%In this section, we aim to develop an automatic pipeline for aligning the given textual commentaries to the video content of corresponding timestamps in existing soccer match datasets.
%In Sec.~\ref{sec:temporal_alignment}, we start by introducing the problem scenario of temporal alignments, Subsequently, in Sec.~\ref{sec:data_preparation}, we introduce the data preparation process, covering manual annotations and data statistics. Finally, we elaborate on our coarse-to-fine multimodal temporal alignment pipeline in Sec.~\ref{sec:coarse_to_fine_temporal_alignment}.

% Finally, we provide an overview and statistics of our meticulously curated {\bf MatchTime} dataset in Section~\ref{sec:dataset_statistics}.

\subsection{Problem Formulation}
\label{sec:temporal_alignment}
Given a soccer match video from the SoccerNet-Caption dataset, {\em i.e.}, $\mathcal{X} = \{\mathcal{V}, \mathcal{C} \}$, where $\mathcal{V} = \{(I_1,\hat{t}_1), \dots, \allowbreak (I_n, \hat{t}_n)\}$ denotes key frames of the video and their corresponding timestamps, and $\mathcal{C} = \{(C_1,t_1), \dots, \allowbreak (C_k,t_k)\}$ represents the $k$ textual commentaries and their provided timestamps in the video, with $n\gg k$.
% and $\mathcal{S} = \{(S_1,\Tilde{t}_1), \dots, (S_j,\Tilde{t}_j)\}$ represents the narration text and their timestamps extracted from the audio with ASR, 
Here, our goal is to improve the soccer game commentary dataset by better aligning textual commentaries with key frames.
Concretely, we adopt a contrastive alignment pipeline to update their timestamps:
$\Tilde{t} = \Phi(\mathcal{V}, \mathcal{C}; \Theta_1)$, 
where $\Theta_1$ denotes the trainable parameters of the alignment model $\Phi$, and $\Tilde{t}$ represents the modified timestamps for all textual commentaries.

\subsection{Method}
\label{sec:coarse_to_fine_temporal_alignment}
As depicted in Figure~\ref{fig:alignment}, 
we propose a two-stage temporal alignment pipeline:
(i) pre-processing with an off-the-shelf automatic speech recognition model~(ASR) and large language model~(LLMs), 
(ii) train an alignment model with contrastive learning. 
We will elaborate on the details as follows.

\vspace{2pt} 
\noindent{\textbf{Pre-processing with ASR and LLMs}.}
We propose to roughly align the textual commentary with video content by leveraging the audio narration, which may include key event descriptions. 
Specifically, we first adopt WhisperX~\cite{bain2022whisperx} for automatic speech recognition (ASR), to obtain the converted narration text with corresponding timestamp intervals from the audio. 
Given that live soccer commentary tends to be fragmented and colloquial, we use LLaMA-3~\cite{llama3} to summarize the ASR results into event descriptions for each 10-second video clip with the prompt described in Appendix~\ref{appendix:event_summarization}.
Subsequently, we feed these event descriptions and the textual commentaries into LLaMA-3 to predict new timestamps for the textual commentaries based on sentence similarities using the prompt detailed in 
Appendix~\ref{appendix:timestamp_prediction}.
\textbf{Note that}, as some videos may not have audio commentary, or narrations that are irrelevant to the video content, such as the background information for certain players, such pre-processing only allows for a coarse-grained alignment of the commentary to video key frames.

\vspace{2pt} 
\noindent{\textbf{Fine-grained Temporal Alignment}.}
Here, we further propose to train a multi-modal temporal alignment model with contrastive learning. 
Concretely, we adopt pre-trained CLIP~\cite{CLIP} to encode textual commentaries and key frames, followed by trainable MLPs, {\em i.e.}, $f(\cdot)$ and $g(\cdot)$: 
\[
\text{C},\text{V} = f(\Phi_{\text{CLIP-T}}(\mathcal{C})), \hspace{3pt} g(\Phi_{\text{CLIP-V}}(\mathcal{V}))
\]
where $\text{C} \in \mathbb{R}^{k \times d}, \text{V} \in \mathbb{R}^{n \times d}$ denotes the resulting textual and visual embeddings, respectively.

We compute the affinity matrix between the textual commentaries and video key frames as:
% \weidi{are the vectors L2 normed? is there temperature in the constrastive loss? }
\[
\hat{\mathbb{A}}{[i, j]} = \frac{\text{C}_i \cdot \text{V}_j}{|| \text{C}_i || \cdot || \text{V}_j ||}, \hspace{5pt} \hat{\mathbb{A}} \in \mathbb{R}^{k \times n}
\]
With the manual annotated {\bf SN-Caption-test-align} as introduced in Sec.~\ref{sec:benchmark_curation}, 
% \weidi{do we bold the dataset name ? or italic?}
we can construct the ground truth label matrix with the same form, {\em i.e.}, $\mathbb{Y} \in \{ 0, 1 \}^{k \times n}$, $\mathbb{Y}{[i, j]} = 1$ if the $i$-th commentary corresponds to the $j$-th key frame, otherwise 0.
% \weidi{I think there is no `keyframe', only `key frame'.}

We train the joint visual-textual embeddings for alignment with contrastive learning~\cite{oord2018infonce}, by maximising similarity scores between the commentary and its corresponding visual frame: 
\begin{align*}
    \mathcal{L}_{\text{align}} = -\frac{1}{k}\sum_{i=1}^{k} \log\bigg[\frac{\sum_{j}^{n} \mathbb{Y}{[i, j]} \exp(\hat{\mathbb{A}}{[i, j]})}{\sum_{j}^{n} \exp(\hat{\mathbb{A}}{[i, j]})}\bigg]
\end{align*}

\vspace{2pt} 
\noindent \textbf{Training and Inference.}
At training time, we use the \textbf{45} manually annotated videos with 2,975 video clip-text pairs from our curated {\bf SN-Caption-test-align}, and leave the \textbf{4} videos for evaluation.
Frames sampled at 1FPS with a two-minute window around the manually annotated ground truth timestamps are utilized for training.
At inference time, considering that data pre-processing has provided a coarse alignment, and there might be replays in soccer match videos, 
we sample frames at 1FPS from 45 seconds before and 30 seconds after the current textual commentary timestamp as visual candidates for alignment.
To validate the effectiveness of our alignment model, 
we evaluate it on 292 samples of 4 unseen annotated matches, 
results can be found in Sec.~\ref{subsec:alignment}.

With the trained model, we perform fine-grained temporal alignment for each textual commentary $\text{C}_i$ by updating its timestamp to  $\Tilde{t}_{i}$ with $\hat{t}_{j}$ of the visual frame $I_j$, which exhibits the highest cross-modal similarity score among all the candidates:
\begin{align*}
    \Tilde{t}_{i} := \hat{t}_{j}, \,\, \mathrm{where} \,\,\, j =  \arg\max(\hat{\mathbb{A}}{[i, :]})
\end{align*}
% where the timestamp of the video frame with the highest matching score is assigned to the input textual commentary.
% We run this alignment model on all training data of SoccerNet-Caption. 
Using the alignment pipeline described above, we have aligned all the pre-processed training data from SoccerNet-Caption.
As for the matches lacking audio, which cannot undergo pre-processing, we directly apply our fine-grained temporal alignment model.
% \weidi{this might be a good reason why people don't train on ASR transcripts, another reason can be, the narrator may includes narrations that are not related to the video content, for example, describe the player's background, number of scores within the entire season, etc.}
As a result, we have aligned 422 videos (373 as the training set and 49 as the validation set), amounting to 29,476 video-text pairs (26,058 for training and 3,418 for validation) in total. 
This contributes a high-quality dataset, termed as {\bf MatchTime}, 
for training an automatic soccer game commentary system.
The detailed statistics of our datasets are listed in Table~\ref{tab:dataset_statistics}.
% \weidi{can we give a table on these ? also indicate which ones are manually annotated, something like Table~\ref{tab:weidi-example}.}

\begin{table}[t]
\begin{center}
\small
\tabcolsep=0.11cm
\begin{tabular}{c c|c c}
% \hline
\toprule
Datasets & Alignment & \# Soccer Matches & \# Commentary \\ 
\midrule
Test & Manual & 49 & 3,267 \\ 
Validation & Auto & 49 & 3,418 \\ 
Training & Auto & 373 & 26,058 \\ 
\bottomrule
\end{tabular}
\end{center}
\vspace{-6pt}
\caption{\textbf{Data Statistics} on our {\bf SN-Caption-test-align} and {\bf MatchTime} datasets.}
\label{tab:dataset_statistics}
\vspace{-0.5cm}
\end{table}

\begin{figure*}[thb]
    \centering
    \includegraphics[width=\textwidth]{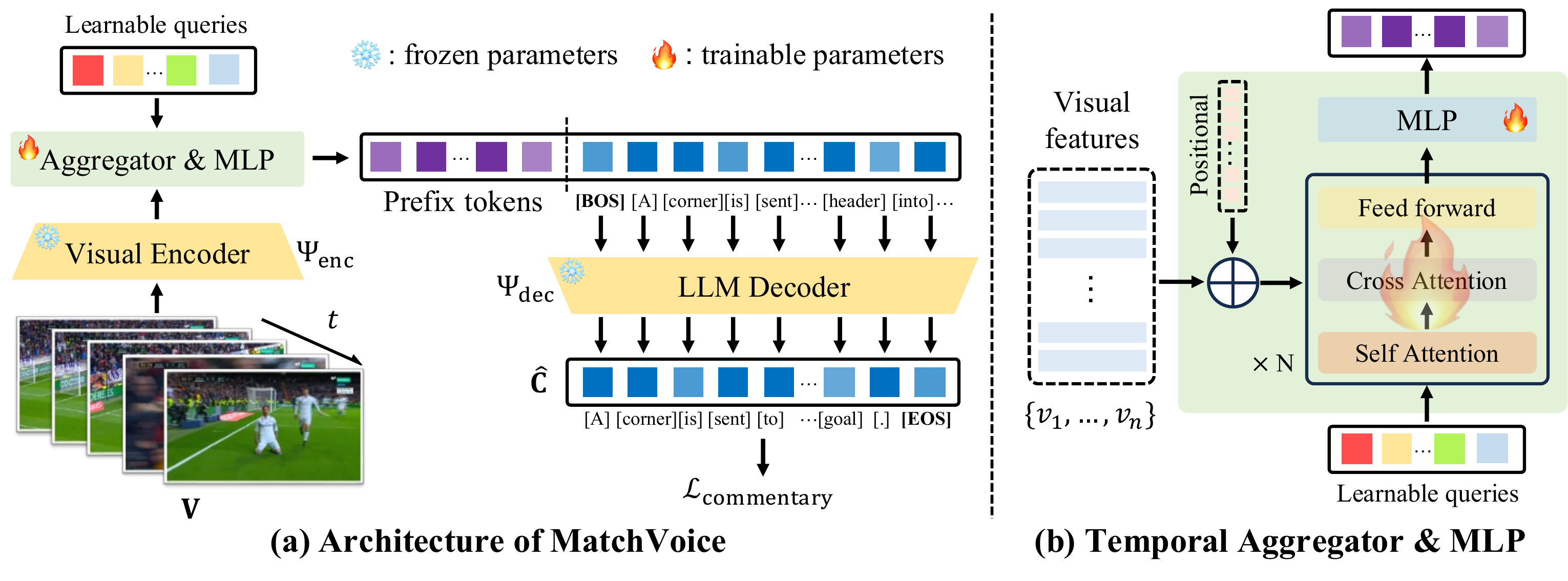} 
    \vspace{-0.8cm}
    \caption{\textbf{MatchVoice Architecture Overview}. Our proposed MatchVoice model leverages a pretrained visual encoder to encode video frames into visual features.
    A learnable temporal aggregator aggregates the temporal information among these features. The temporally aggregated features are then projected into prefix tokens of LLM via a trainable MLP projection layer, to generate the corresponding textual commentary.}
    \label{fig:commentary}
    \vspace{-0.3cm}
\end{figure*}

\section{Automatic Soccer Game Commentary}
\label{sec:automatic_soccer_game_commentary}
Based on the curated dataset, we consider training a visual-language model for automatic commentary generation on given input video segments, termed as \textbf{MatchVoice}. Specifically, we start by describing the problem scenario, and followed by detailing on our proposed architecture.

\vspace{2pt}
\noindent{\textbf{Problem Formulation}.}
Given a soccer game video with multiple clips, {\em i.e.}, $\mathcal{V} = \{\mathbf{V}_1, \mathbf{V}_2, \dots, \mathbf{V}_T\}$, our goal is to develop a visual-language model that generates corresponding textual commentary for each video segment, {\em i.e.}, $\hat{\mathbf{C}}_i = \Psi(\mathbf{V}_i; \Theta_2)$, where $\Theta_2$ refers to the trainable parameters.

\vspace{3pt} \noindent{\textbf{Architecture.}}
As depicted in Figure~\ref{fig:commentary}, our proposed model comprises of three components. 
Here, we focus on processing one segment, and ignore the subscripts for simplicity.

{\em First}, we adopt the frozen, pre-trained visual encoder to compute the framewise features within the video clip, {\em i.e.}, $\{v_1, v_2, \dots, v_n\} = \Psi_{\text{enc}}(\mathbf{V})$.
Note that, all visual encoders are framewise, except InternVideo, which takes 8 frames per second and aggregates them into 1 feature vector by itself.
% \weidi{into one feature vector or one frame ?}

%feature extraction, each video segment $\mathbf{V} = \{ \mathbf{I}_{i,1}, \mathbf{I}_{i,2}, \dots, \mathbf{I}_{i,n} \}$ is encoded into visual features $\{\mathbf{F}_{i,1}, \mathbf{F}_{i,2}, \dots, \mathbf{F}_{i,n} \}$ by a frozen visual encoder $\Psi_{enc}$.

{\em Second}, we use a Perceiver-like architecture~\cite{jaegle2021perceiver} aggregator to aggregate the temporal information among visual features. 
Specifically, we adopt two transformer decoder layers, with a fixed-length learnable query, and visual features as keys and values, to obtain the temporally-aware features, {\em i.e.}, $\mathbf{F} = \Psi_{\text{agg}}(v_1, v_2, \dots, v_n)$.  

{\em Last}, an MLP projection layer is used to map the output queries into desired feature dimensions, used as prefix tokens for a decoder-only large language model~(LLMs), to generate the desired textual commentary, 
{\em i.e.}, $\hat{\mathbf{C}} = \Psi_{\text{dec}}(\Psi_{\text{proj}}(\mathbf{F}))$.
With the ground truth commentary for the soccer video clips, 
the model is then trained with standard negative log-likelihood loss for language generation.

% \weidi{if you want to give abstract math, use the notations from previous paragraph, otherwise delete the notations in previous par.}
% \weidi{same, decide whether to use the following notations, if yes, 
% check them to show they have indeed reflected the operation you have done.}
% \begin{align*}
%     \hat{\mathbf{C}} = \Psi_{dec}({\mathrm{prefix}})
% \end{align*}

%Negative Log-Likelihood (NLL) loss as:
%\begin{align*}
%    \mathcal{L}_{\text{commentary}} = -\mathrm{log}\ p_{\Theta_2}\ (\mathbf{C}|\mathbf{V})
%\end{align*}
%By minimizing such NLL loss, we can train the commentary model by prompt-tuning, with the image encoder and LLM decoder frozen.

%\begin{align*}
%    \mathbf{F}_{i,j} & = \Psi_{enc}(\mathbf{I}_{i,j}), \  1 \leq i \leq T,\  1 \leq j \leq n \\
%    {\mathrm{prefix}}_{i} & = \mathrm{proj}(\mathrm{QFormer}(\mathbf{F}_{i,1}, \mathbf{F}_{i,2}, \dots, \mathbf{F}_{i,n}))
%\end{align*}

% For each video segment $\mathbf{V}_i$, the frames are first extracted with 1 or 2 FPS to $\{ \mathbf{I}_{i,1}, \mathbf{I}_{i,2}, \dots, \mathbf{I}_{i,n} \}$, and encoded by the visual encoder $\Psi_{enc}$ to features $\{\mathbf{F}_{i,1}, \mathbf{F}_{i,2}, \dots, \mathbf{F}_{i,n} \}$. 

% Finally, we input these prefix tokens, which contain the visual features, into the language decoder model to synthesize the final commentary $\mathbf{y}_i$:
% By reducing such NLL loss, we train the soccer game commentary model by prompt-tuning method. 
% This approach allows us to finalize the commentary while keeping both the image encoder and the language decoder frozen.

\begin{table}[t]
    \centering
        
    % \footnotesize
    \small
    \tabcolsep=0.24cm
     \begin{tabular}{c|cccc}
     \toprule
    \multicolumn{1}{c|}{Pre-processing} & \XSolidBrush & \Checkmark & \XSolidBrush & \Checkmark \\
    \multicolumn{1}{c|}{Contrastive-Align} & \XSolidBrush & \XSolidBrush & \Checkmark & \Checkmark \\
          \midrule
  $\mathrm{avg} (\Delta)$ (s)        & 10.21 & -0.96 & 6.35 & \textbf{0.03} \\
  $\mathrm{avg} (| \Delta |)$ (s)    & 13.89 & 13.75 & 12.15 & \textbf{6.89} \\
  \midrule
  $\mathrm{window}_{10}$ (\%)        & 35.32 & 34.86 & 77.06 & \textbf{80.73} \\
  $\mathrm{window}_{30}$ (\%)        & 65.60 & 69.72 & 83.49 & \textbf{91.28} \\ 
  $\mathrm{window}_{45}$ (\%)        & 77.98 & 80.28 & 86.70 & \textbf{95.41} \\ 
  $\mathrm{window}_{60}$ (\%)        & 88.07 & 85.32 & 90.37 & \textbf{98.17} \\ 

  \bottomrule
        \end{tabular}
        \vspace{-2pt}
        \caption{\textbf{Alignment Statistics.} 
        We report the temporal offset statistics on 4 manually annotated test videos (comprising a total of 292 samples).
        $\Delta$ and $\mathrm{window}_{t}$ represent the temporal offset and the percentage of commentaries that fall within a window of $t$ seconds around the visual key frames, respectively.
        }
        \vspace{-0.2cm}
        \label{tab:alignment_quantitative_results}
\end{table}

\section{Experiments}
In this section, we separately describe the experiment results for the considered tasks, namely, soccer commentary alignment~(Sec.~\ref{subsec:alignment}), 
and automatic soccer commentary generation~(Sec.~\ref{subsec:generation}).

\begin{table*}[t]
\begin{center}
\footnotesize
\tabcolsep=0.26cm
\begin{tabular}{cccccccc}
\toprule

\multicolumn{1}{c|}{Method} & \multicolumn{1}{c|}{Visual Features} &  \multicolumn{1}{c|}{BLEU-1} & \multicolumn{1}{c|}{BLEU-4} & \multicolumn{1}{c|}{METEOR} & \multicolumn{1}{c|}{ROUGE-L} & \multicolumn{1}{c|}{CIDEr} & \multicolumn{1}{c}{GPT-score} \\ 

% NEW
% \midrule
% \multicolumn{8}{c}{Zero-shot} \\ 
% \midrule
% \multicolumn{1}{c|}{\multirow{2}{*}{Video-LLaMA} & \multicolumn{1}{c|}{Video-LLaMA(7B)} & \multicolumn{1}{c|}{26.81} & \multicolumn{1}{c|}{5.24} & \multicolumn{1}{c|}{23.57} & \multicolumn{1}{c|}{23.12} & \multicolumn{1}{c|}{13.78} & \multicolumn{1}{c}{6.27} \\ 
% \multicolumn{1}{c|}{} & \multicolumn{1}{c|}{Video-LLaMA(13B)} & \multicolumn{1}{c|}{\textcolor{blue}{\underline{29.74}}} & \multicolumn{1}{c|}{\textcolor{blue}{\underline{7.31}}} &  \multicolumn{1}{c|}{\textcolor{red}{\textbf{26.40}}} & \multicolumn{1}{c|}{26.19} & \multicolumn{1}{c|}{23.74} & \multicolumn{1}{c}{6.84} \\ 
\midrule
\multicolumn{8}{c}{Zero-shot} \\ 
\midrule
\multicolumn{1}{c|}{\multirow{1}{*}{Video-LLaMA(7B)}} & \multicolumn{1}{c|}{ViT} & \multicolumn{1}{c|}{\textcolor{red}{\textbf{12.95}}} & \multicolumn{1}{c|}{\textcolor{blue}{\underline{0.52}}} & \multicolumn{1}{c|}{\textcolor{blue}{\underline{6.11}}} & \multicolumn{1}{c|}{\textcolor{blue}{\underline{15.06}}} & \multicolumn{1}{c|}{\textcolor{red}{\textbf{1.97}}} & \multicolumn{1}{c}{\textcolor{blue}{\underline{2.91}}} \\ 
% \cline{2-2}
% \midrule
\multicolumn{1}{c|}{Video-LLaMA(13B)} & \multicolumn{1}{c|}{ViT} & \multicolumn{1}{c|}{\textcolor{blue}{\underline{12.64}}} & \multicolumn{1}{c|}{\textcolor{red}{\textbf{0.58}}} & \multicolumn{1}{c|}{\textcolor{red}{\textbf{6.75}}} &  \multicolumn{1}{c|}{\textcolor{red}{\textbf{20.47}}} & \multicolumn{1}{c|}{\textcolor{blue}{\underline{1.76}}} & \multicolumn{1}{c}{\textcolor{red}{\textbf{3.78}}} \\  
% NEW

\midrule
\multicolumn{8}{c}{Trained on original SoccerNet-Caption} \\ 
\midrule
\multicolumn{1}{c|}{\multirow{3}{*}{SN-Caption}} & \multicolumn{1}{c|}{C3D} & \multicolumn{1}{c|}{22.13} & \multicolumn{1}{c|}{4.25} & \multicolumn{1}{c|}{23.14} & \multicolumn{1}{c|}{23.25} & \multicolumn{1}{c|}{11.97} & \multicolumn{1}{c}{5.80} \\ \cline{2-2}
                        \multicolumn{1}{c|}{} & \multicolumn{1}{c|}{ResNet} & \multicolumn{1}{c|}{26.46} & \multicolumn{1}{c|}{5.33} & \multicolumn{1}{c|}{23.58} &  \multicolumn{1}{c|}{23.58} & \multicolumn{1}{c|}{13.71} & \multicolumn{1}{c}{6.28} \\ \cline{2-2}
                        \multicolumn{1}{c|}{} & \multicolumn{1}{c|}{Baidu} & \multicolumn{1}{c|}{\textcolor{blue}{\underline{29.61}}} & \multicolumn{1}{c|}{\textcolor{blue}{\underline{6.83}}} &  \multicolumn{1}{c|}{\textcolor{red}{\textbf{25.38}}} & \multicolumn{1}{c|}{25.28} & \multicolumn{1}{c|}{20.61} & \multicolumn{1}{c}{6.72}  \\ 

\midrule
\multicolumn{1}{c|}{\multirow{5}{*}{\begin{tabular}[c]{@{}c@{}}\textbf{MatchVoice}\\ (\textbf{Ours})\end{tabular}}} & \multicolumn{1}{c|}{C3D} & \multicolumn{1}{c|}{28.85} & \multicolumn{1}{c|}{5.62} & \multicolumn{1}{c|}{23.29} &  \multicolumn{1}{c|}{26.69} & \multicolumn{1}{c|}{19.06} & \multicolumn{1}{c}{\textcolor{blue}{\underline{6.90}}} \\ \cline{2-2}
            \multicolumn{1}{c|}{} & \multicolumn{1}{c|}{ResNet} & \multicolumn{1}{c|}{28.75} & \multicolumn{1}{c|}{5.87} & \multicolumn{1}{c|}{23.78} &  \multicolumn{1}{c|}{26.69} & \multicolumn{1}{c|}{20.65} & \multicolumn{1}{c}{6.75} \\ \cline{2-2}
            \multicolumn{1}{c|}{} & \multicolumn{1}{c|}{InternVideo} & \multicolumn{1}{c|}{\multirow{1}{*}{28.50}} & \multicolumn{1}{c|}{6.24} & \multicolumn{1}{c|}{24.30} & \multicolumn{1}{c|}{\textcolor{red}{\textbf{30.75}}}  & \multicolumn{1}{c|}{23.34} & \multicolumn{1}{c}{6.80} \\ \cline{2-2}
            \multicolumn{1}{c|}{} & \multicolumn{1}{c|}{CLIP} & \multicolumn{1}{c|}{28.65} & \multicolumn{1}{c|}{6.62} &  \multicolumn{1}{c|}{24.20} & \multicolumn{1}{c|}{27.33} & \multicolumn{1}{c|}{\textcolor{blue}{\underline{24.35}}} & \multicolumn{1}{c}{6.78} \\ \cline{2-2}
            \multicolumn{1}{c|}{} & \multicolumn{1}{c|}{Baidu} & \multicolumn{1}{c|}{\textcolor{red}{\textbf{30.32}}} & \multicolumn{1}{c|}{\textcolor{red}{\textbf{8.45}}} &  \multicolumn{1}{c|}{\textcolor{blue}{\underline{25.25}}} & \multicolumn{1}{c|}{\textcolor{blue}{\underline{29.40}}} & \multicolumn{1}{c|}{\textcolor{red}{\textbf{33.84}}} & \multicolumn{1}{c}{\textcolor{red}{\textbf{7.07}}} \\  
\midrule
\multicolumn{8}{c}{Trained on our aligned MatchTime} \\ 

\midrule
\multicolumn{1}{c|}{\multirow{3}{*}{SN-Caption}} & \multicolumn{1}{c|}{C3D} & \multicolumn{1}{c|}{26.81} & \multicolumn{1}{c|}{5.24} & \multicolumn{1}{c|}{23.57} & \multicolumn{1}{c|}{23.12} & \multicolumn{1}{c|}{13.78} & \multicolumn{1}{c}{6.27} \\ \cline{2-2}
                        \multicolumn{1}{c|}{} & \multicolumn{1}{c|}{ResNet} & \multicolumn{1}{c|}{27.63} & \multicolumn{1}{c|}{5.75} &  \multicolumn{1}{c|}{24.05} & \multicolumn{1}{c|}{23.42} & \multicolumn{1}{c|}{15.65} & \multicolumn{1}{c}{6.33} \\ \cline{2-2}
                        \multicolumn{1}{c|}{} & \multicolumn{1}{c|}{Baidu} & \multicolumn{1}{c|}{\textcolor{blue}{\underline{29.74}}} & \multicolumn{1}{c|}{\textcolor{blue}{\underline{7.31}}} &  \multicolumn{1}{c|}{\textcolor{red}{\textbf{26.40}}} & \multicolumn{1}{c|}{26.19} & \multicolumn{1}{c|}{23.74} & \multicolumn{1}{c}{6.84} \\ 
\midrule
\multicolumn{1}{c|}{\multirow{5}{*}{\begin{tabular}[c]{@{}c@{}}\textbf{MatchVoice}\\ (\textbf{Ours})\end{tabular}}} & \multicolumn{1}{c|}{C3D} & \multicolumn{1}{c|}{28.67} & \multicolumn{1}{c|}{6.55} & \multicolumn{1}{c|}{24.46} &  \multicolumn{1}{c|}{27.38} & \multicolumn{1}{c|}{26.53} & \multicolumn{1}{c}{6.89}  \\ \cline{2-2}
            \multicolumn{1}{c|}{} & \multicolumn{1}{c|}{ResNet} & \multicolumn{1}{c|}{29.21} & \multicolumn{1}{c|}{6.60} & \multicolumn{1}{c|}{24.11} &  \multicolumn{1}{c|}{24.32} & \multicolumn{1}{c|}{28.56} & \multicolumn{1}{c}{6.84} \\ \cline{2-2}
            \multicolumn{1}{c|}{} & \multicolumn{1}{c|}{InternVideo} & \multicolumn{1}{c|}{\multirow{1}{*}{29.18}} & \multicolumn{1}{c|}{6.89} & \multicolumn{1}{c|}{25.04} & \multicolumn{1}{c|}{28.18}  & \multicolumn{1}{c|}{\textcolor{blue}{\underline{30.22}}} & \multicolumn{1}{c}{\textcolor{blue}{\underline{6.99}}} \\ \cline{2-2}
            \multicolumn{1}{c|}{} & \multicolumn{1}{c|}{CLIP} & \multicolumn{1}{c|}{29.56} & \multicolumn{1}{c|}{6.90} &  \multicolumn{1}{c|}{24.62} & \multicolumn{1}{c|}{\textcolor{red}{\textbf{31.25}}} & \multicolumn{1}{c|}{28.66} & \multicolumn{1}{c}{6.82}  \\ \cline{2-2}
            \multicolumn{1}{c|}{} & \multicolumn{1}{c|}{Baidu} & \multicolumn{1}{c|}{\textcolor{red}{\textbf{31.42}}} & \multicolumn{1}{c|}{\textcolor{red}{\textbf{8.92}}} &  \multicolumn{1}{c|}{\textcolor{blue}{\underline{26.12}}} & \multicolumn{1}{c|}{\textcolor{blue}{\underline{29.66}}} & \multicolumn{1}{c|}{\textcolor{red}{\textbf{38.42}}} & \multicolumn{1}{c}{\textcolor{red}{\textbf{7.08}}} \\ 

\midrule
\multicolumn{8}{c}{Apply LoRA to the LLM decoder in MatchVoice} \\ 
\midrule

\multicolumn{1}{c|}{\multirow{1}{*}{Frozen LLM}} & \multicolumn{1}{c|}{Baidu} & \multicolumn{1}{c|}{31.42} & \multicolumn{1}{c|}{8.92} & \multicolumn{1}{c|}{26.12} & \multicolumn{1}{c|}{\textcolor{red}{\textbf{29.66}}} & \multicolumn{1}{c|}{38.42} & \multicolumn{1}{c}{7.08} \\ 
\multicolumn{1}{c|}{\multirow{1}{*}{Rank = 8\hspace{0.15cm}}} & \multicolumn{1}{c|}{Baidu} & \multicolumn{1}{c|}{30.85} & \multicolumn{1}{c|}{8.77} & \multicolumn{1}{c|}{26.45} & \multicolumn{1}{c|}{\textcolor{blue}{\underline{26.44}}} & \multicolumn{1}{c|}{37.72} & \multicolumn{1}{c}{7.21} \\ 
% \cline{2-2}
% \midrule
\multicolumn{1}{c|}{Rank = 16} & \multicolumn{1}{c|}{Baidu} & \multicolumn{1}{c|}{\textcolor{red}{\textbf{33.22}}} & \multicolumn{1}{c|}{\textcolor{red}{\textbf{10.10}}} & \multicolumn{1}{c|}{\textcolor{red}{\textbf{26.79}}} &  \multicolumn{1}{c|}{26.06} & \multicolumn{1}{c|}{\textcolor{blue}{\underline{39.27}}} & \multicolumn{1}{c}{\textcolor{blue}{\underline{7.32}}} \\ 
\multicolumn{1}{c|}{\multirow{1}{*}{Rank = 32}} & \multicolumn{1}{c|}{Baidu} & \multicolumn{1}{c|}{\textcolor{blue}{\underline{31.55}}} & \multicolumn{1}{c|}{\textcolor{blue}{\underline{9.33}}} & \multicolumn{1}{c|}{\textcolor{blue}{\underline{26.53}}} & \multicolumn{1}{c|}{21.62} & \multicolumn{1}{c|}{\textcolor{red}{\textbf{42.00}}} & \multicolumn{1}{c}{7.23} \\ 
% \cline{2-2}
% \midrule
\multicolumn{1}{c|}{Rank = 64} & \multicolumn{1}{c|}{Baidu} & \multicolumn{1}{c|}{30.71} & \multicolumn{1}{c|}{8.63} & \multicolumn{1}{c|}{26.36} &  \multicolumn{1}{c|}{24.32} & \multicolumn{1}{c|}{35.33} & \multicolumn{1}{c}{\textcolor{red}{\textbf{7.35}}} \\ 
% \midrule

\bottomrule
\end{tabular}

    \vspace{-2pt}
    \caption{\textbf{Quantitative Comparison on Commentary Generation}. 
     All variants of SN-caption baseline methods, our MatchVoice are retrained on both the original unaligned SoccerNet-Caption and our temporally aligned MatchTime training sets, while MatchVoice with LoRA applied on LLM decoder was trained on MatchTime training sets for only.
     All the commentary models are evaluated on our manually curated SN-Caption-test-align benchmark. 
    In each unit, we denote the best performance in \textcolor{red}{\textbf{RED}} and the second-best performance in \textcolor{blue}{\underline{$\mathrm{BLUE}$}}.
% \textcolor{red}{\textbf{RED}}: best performance, \textcolor{blue}{\underline{BLUE}}: second best performance.
}
\label{tab:commentary_quantitative_results}
\vspace{0.2cm}
\end{center}
\end{table*}

\subsection{Video-Commentary Temporal Alignment}
\label{subsec:alignment}
In this part, we first introduce the implementation details and evaluation metrics of our temporal alignment pipeline, followed by a quantitative comparison and analysis of the alignment results.
% \weidi{Give one summary sentence.}

\vspace{2pt}
\noindent{\textbf{Implementation Details.}}
We use pretrained off-the-shelf CLIP ViT-B/32 model to extract visual and textual features for our alignment pipeline, which are then passed through two MLP layers to get 512-dim features for contrastive learning. We use the AdamW~\cite{adamW} optimizer and the learning rate is set to $5 \times 10^{-4}$ to train the alignment model for 50 epochs.

\vspace{2pt} \noindent{\textbf{Evaluation Metrics.}}
To evaluate temporal video-text alignment quality, 
we report various metrics on 4 unseen videos (with 292 samples) from our curated {\bf SN-Caption-test-align} benchmark, including the average temporal offset ($\mathrm{avg}(\Delta)$), the average absolute temporal offset ($\mathrm{avg}(| \Delta |)$), and the percentage of textual commentaries falling within 10s, 30s, 45s, and 60s windows around each key frame.

\vspace{2pt}
\noindent{\textbf{Quantitative Results.}}
As depicted in Table~\ref{tab:alignment_quantitative_results}, our proposed automatic temporal alignment pipeline effectively aligns visual content and textual commentary in a coarse-to-fine manner.
Specifically, our approach reduces the average absolute offset by \textbf{7.0s} (from 13.89 seconds to 6.89 seconds) and significantly enhances the alignment of textual commentary with key frames. 
% \textcolor{edit}{Relevant ablation experiments in \ref{appendix:temporal_alignment} indicate that our alignment model represents its best performance when using a 120-second window. Also,} 
It is important to highlight that, in comparison to solely using a contrastive alignment model, incorporating data pre-processing enhances coarse alignment. This provides a robust foundation for subsequent fine-grained alignment, consistently leading to further improvements in performance. Furthermore, the proportion of commentary that aligns within a precise 10-second window increases dramatically by \textbf{45.41\%} (from 35.32\% to 80.73\%). Remarkably, nearly all (\textbf{98.17\%}) textual commentaries now fall within a 60-second window surrounding the key frames, underscoring the efficacy of our two-stage alignment pipeline.

\subsection{Soccer Commentary Generation}
\label{subsec:generation}
In this part, we first detail on the implementation details and evaluation metrics of the commentary generation model.
Then, we analyze the results from both quantitative and qualitative perspectives.
Finally, we validate the effectiveness of the modules through ablation experiments.

\vspace{2pt}
\noindent{\textbf{Implementation Details.}}
Our automatic commentary model can employ various visual features such as C3D~\cite{c3d}, ResNet~\cite{he2016resnet}, Baidu~\cite{baidu}, CLIP~\cite{CLIP}, and InternVideo~\cite{wang2022internvideo}. All visual features are extracted from the video at 2FPS, except for InternVideo and Baidu, which are extracted at 1FPS.
% The query length of the temporal aggregator is fixed at 32, 
The number of query vectors in the temporal aggregator is fixed at 32, 
% \weidi{you mean the number of query vectors ?} 
and the MLP projection layer projects the aggregated features to a 768-dimensional prefix token that is then fed into LLaMA-3~\cite{llama3} for decoding the textual commentaries. The learning rate is set to $1 \times 10^{-4}$ to train the commentary model for 100 epochs. All experiments are conducted with one single Nvidia RTX A100 GPU. For baselines, we retrain several variants of SN-Caption~\cite{densecap} using its official implementation.
% NetVLAD++~\cite{netvlad} is adopted to aggregate the temporal information of the extracted features.
% Then the pooled features are decoded by an LSTM~\cite{hochreiter1997long}.
% \haoning{Above descriptions on baselines can be in supplementary...}

% For alignment, 45 manually annotated matches (with 2,975 samples) are used to train the contrastive model to temporally align the noisy training set.
% The commentary model is then trained using all automatically aligned videos of the training set.

% During inference, we sample frames at 1FPS from 45 seconds before to 30 seconds after the current anchor timestamp for fine-grained alignment.
% The timestamp of the video frame with the highest matching score is assigned to the input textual commentary.
% Additionally, for the matches that lack audio and thus cannot undergo coarse alignment, we directly apply contrastive temporal alignment.

% We use the AdamW~\cite{adamW} optimizer and the learning rate is set to $5 \times 10^{-4}$ to train the alignment model for 50 epochs, 
% \begin{figure*}[t]
%   \centering
%   \includegraphics[width=\textwidth]{latex/images/alignment_qualative.pdf} \\
%   \vspace{-2pt}
%   \caption{\textbf{Qualitative results on Temporal Alignment.} Timestamps before alignment are in \textcolor{orange}{Orange}, those after alignment are in \textcolor[rgb]{0,0.5,0}{Green}.}
%     \vspace{-6pt}
%  \label{fig:alignment_qualitative_results}
% \end{figure*}

\vspace{4pt}
\noindent{\textbf{Evaluation Metrics.}}
To evaluate the quality of generated textual commentaries, 
we adopt various popular metrics, including BLEU~(B)~\cite{papineni2002bleu}, METEOR~(M)~\cite{banerjee2005meteor}, ROUGE-L~(R-L)~\cite{lin2004rouge}, CIDEr~(C)~\cite{vedantam2015cider}.
Additionally, we also report the GPT-score~\cite{fu2023gptscore}, ranging from 1 to 10, based on semantic information, expression accuracy, and professionalism. 
This score is provided by GPT-3.5 using the ground truth and generated textual commentary as inputs, with the prompt described in Appendix~\ref{appendix:evaluation_metrics}.
% Additionally, we also report the GPT-score~\cite{fu2023gptscore} ranging from 1 to 10 in terms of semantic information, expression accuracy, and professionalism, given by GPT-3.5 with the ground truth and generated textual commentary as inputs, using the prompt described in Appendix~\ref{appendix:evaluation_metrics}. 
% \weidi{give reference for the GPT-score.}
% \weidi{give the prompt in appendix, and refer to it.}

\vspace{4pt}
\noindent{\textbf{Quantitative Results.}}
As depicted in Table~\ref{tab:commentary_quantitative_results}, we can draw the following four observations:
(i) Off-the-shelf vision-language models struggle to achieve satisfactory performance on the soccer game commentary generation task in a zero-shot manner, indicating that the professional nature of this task requires additional training on specific data to be adequately addressed;
(ii) Our proposed {\bf MatchVoice} significantly outperforms existing methods in generating professional soccer game commentary, establishing new state-of-the-art performance;
(iii) Both the baseline methods and our {\bf MatchVoice} benefit from temporally aligned data, demonstrating the superiority and necessity of temporal alignment;
(iv) Commentary models based on Baidu visual encoder perform better than others, we conjecture this is because the pretraining on soccer data further improves the quality of commentary generation.

\begin{figure*}[t]
  \centering
  \includegraphics[width=\textwidth]{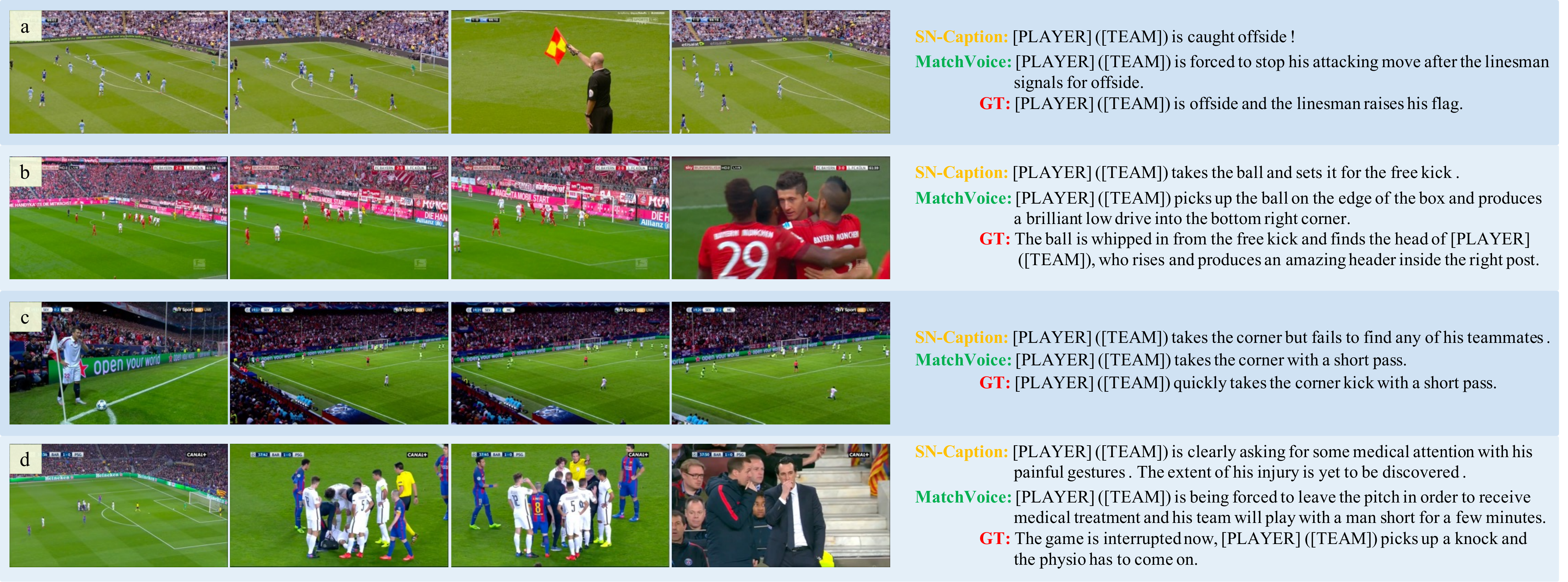} \\
  \vspace{-2pt}
  \caption{\textbf{Qualitative results on commentary generation.}
  Our MatchVoice demonstrates advantages in multiple aspects: (a) richer semantic descriptions, (b) full commentaries of multiple incidents in a single video, (c) accuracy of descriptions, and (d) predictions of incoming events.
  }
    \vspace{-6pt}
 \label{fig:qualitative_results}
\end{figure*}

\begin{table}[t]
\begin{center}
\small
\tabcolsep=0.17cm

\begin{tabular}{c|c|c|c|c|c|c}
\toprule
Align & Win~(s) & B@1 & B@4 & M & R-L & C \\ 
\midrule
\multicolumn{1}{c|}{\multirow{4}{*}{\XSolidBrush}} & 10 & 25.02 & 5.00 & 23.32 & 24.65 & 19.34 \\ 
 & 30 &  \underline{30.32} & 8.45 & 25.25 & \underline{29.40} & 33.84 \\ 
 & 45 &  30.29 & 7.97 & 25.26 & 24.62 & 29.37 \\ 
 & 60 &  30.08 & \underline{8.60} & \underline{25.41} & 23.96 & \underline{35.08} \\ 
\midrule
\multicolumn{1}{c|}{\multirow{4}{*}{\Checkmark}} & 10 & 29.01 & 8.38 & 25.49 & 24.94 & \textbf{40.51} \\ 
 & 30 &  \textbf{31.42} & \textbf{8.92} & \textbf{26.12} & \textbf{29.66} & 38.42 \\ 
 & 45 &  30.07 & 8.32 & 25.65 & 29.65 & 36.51 \\ 
 & 60 &  29.87 & 8.13 & 25.43 & 24.30 & 36.00 \\
\bottomrule
\end{tabular}
\end{center}
\vspace{-6pt}
\caption{\textbf{Ablation study on window size}. 
Using the visual content within 30s around key frames yields the best commentary performance, and temporal alignment of data leads to a universal performance improvement.
}

\label{tab:ablation_studies_window_size}
\vspace{-0.2cm}
\end{table}

\begin{figure*}[t]
  \centering
  \includegraphics[width=\textwidth]{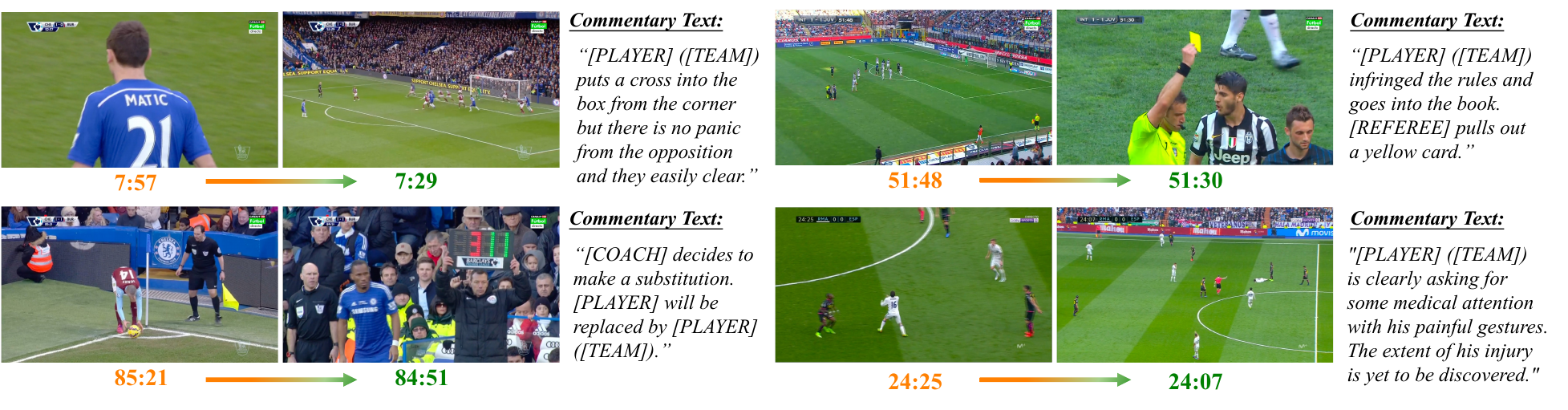} \\
  % \vspace{-2pt}
  \caption{\textbf{Qualitative results on Temporal Alignment.} For the same commentary text, original timestamps in SoccerNet-Caption are in \textcolor{orange}{Orange}, those timestamps after alignment in MatchTime are in \textcolor[rgb]{0,0.5,0}{Green}.}
    \vspace{-6pt}
 \label{fig:alignment_qualitative_results}
\end{figure*}

\vspace{4pt} \noindent{\textbf{Qualitative Results.}}
In Figure~\ref{fig:alignment_qualitative_results}, 
we present qualitative examples on temporal alignment, 
showing that our model enables to correctly align the commentary text with corresponding visual frame. 
In Figure~\ref{fig:qualitative_results},
we show the predictions from our {\bf MatchVoice} model, 
and compare them with baseline results and ground truth. 
It can be seen that our proposed model can generate accurate textual commentaries for professional soccer games that are rich in semantic information.

\vspace{3pt} 
\noindent{\textbf{Ablation Studies.}}
All ablation experiments are conducted using MatchVoice with Baidu features.

\textbf{(i) Window Size.}
The size of the temporal window affects the number of input frames, which in turn impacts the performance of commentary generation.
Therefore, we sample frames within windows of 10s, 30s, 45s, and 60s around the commentary timestamps, and then train and evaluate the commentary generation model to assess the effect of window size on generation quality.
As shown in Table~\ref{tab:ablation_studies_window_size}, 
our {\bf MatchVoice} performs best with a window size of 30 seconds, which is shorter than the 45s window raised in previous work~\cite{densecap}. 
This indicates that our alignment pipeline precisely synchronizes visual information with the corresponding timestamps.
Additionally, the aligned data improves performance across all temporal window settings, especially in the extreme case of a 10s window, demonstrating the necessity of temporal alignment.

\begin{table}[t]
\begin{center}
\small
\tabcolsep=0.16cm
\begin{tabular}{cc|c|c|c|c|c}
% \hline
\toprule
Coarse & Fine & B@1 & B@4 & M & R-L & C \\ 
\midrule
\XSolidBrush & \XSolidBrush &  30.32 & 8.45 & 25.25 & 29.40 & 33.84  \\ 
\Checkmark   & \XSolidBrush &  30.52 & 8.90 & 25.73 & 28.18 & 37.53  \\ 
\XSolidBrush & \Checkmark   &  30.55 & 8.81 & 26.03 & 29.40 & 36.13  \\ 
\Checkmark   & \Checkmark   &  \textbf{31.42} & \textbf{8.92} & \textbf{26.12} & \textbf{29.66} & \textbf{38.42} \\ 
\bottomrule
\end{tabular}
\end{center}
\vspace{-6pt}
\caption{\textbf{Ablation study on alignment strategy}.
    The quality of temporal alignment is directly reflected in downstream commentary generation tasks: better alignment leads to better commentary generation quality.
    }
\label{tab:ablation_studies_alignment_strategy}
\vspace{-0.2cm}
\end{table}

\textbf{(ii) Alignment Strategy.}
To validate the benefits of temporal alignment on downstream tasks, we train our MatchVoice model using data with different levels of alignment, with a fixed window size of 30 seconds, and compare their performance 
(where `Coarse' refers to only data pre-processing 
and `Fine' stands for fine-grained temporal alignment).
As depicted in Table~\ref{tab:ablation_studies_alignment_strategy}, 
compared to using the original misaligned dataset, training on either coarse-aligned or fine-aligned data significantly improves performance.
Furthermore, the model trained on the two-stage aligned data exhibits the largest performance improvement, which demonstrates the necessity of temporal alignment to boost commentary generation quality.
% As depicted in Table~\ref{tab:ablation_studies_alignment_strategy}, 
% compared to using the original misaligned dataset, the models trained on our automatically aligned and filtered dataset demonstrate superior commentary generation performance. 
% This highlights the necessity of temporal alignment to boost commentary generation quality.

\textbf{(iii) LoRA on LLMs Decoder.}
Given that the Baidu visual encoder pretrained on soccer data could potentially boost performance, we further investigate the impact of fine-tuning the language decoder on soccer-specific data.
Considering the high computational cost of fine-tuning the entire LLM, we introduce a small number of trainable LoRA~\cite{hu2021lora} layers within the LLMs decoder to capture the priors from soccer game commentary data.
As presented in Table~\ref{tab:commentary_quantitative_results}, introducing these LoRA layers leads to notable performance improvements, highlighting the necessity of leveraging soccer-specific priors within the dataset.

\section{Related Works}

% \noindent{\textbf{Temporal Video-Text Alignment.}}
\noindent{\textbf{Temporal video-text alignment}} aims to precisely associate textual descriptions or narratives with their corresponding video segments.
Large-scale instructional videos such as HowTo100M~\cite{miech2019howto100m} and YouCook2~\cite{zhou2018youcook2} have already catalyzed extensive multi-modal alignment works based on vision-language co-training.
Concretely, TAN~\cite{han2022temporal} directly aligns procedure narrations transcribed through Automatic Speech Recognition~(ASR) with video segments.
DistantSup~\cite{lin2022learning} and VINA~\cite{mavroudi2023learning} further explore leveraging external knowledge bases~\cite{koupaee2018wikihow} to assist the alignment process, while \citet{li2023strong} propose integrating both action and step textual information to accomplish the video-text alignment.

In this paper, we train a multi-modal alignment model to automatically correct existing data and build a higher-quality soccer game commentary dataset.
Moreover, we further demonstrate the superiority and indispensability of our alignment pipeline through downstream commentary tasks, confirming its critical significance.
% In this paper, we extend temporal video-text alignment to the domain of soccer, utilizing video, audio, and textual commentary data to perform automatic alignment in a coarse-to-fine manner.
% We demonstrate the superiority and indispensability of our alignment pipeline through downstream commentary tasks, confirming its critical significance.

\vspace{2pt}
% \noindent{\textbf{Image \& Video Captioning.}}
\noindent{\textbf{Video captioning}} has been a long-standing research challenge in computer vision~\cite{krishna2017dense, yang2023vid2seq}, primarily due to the limited annotation and expensive computation. Benefiting from the development of LLMs, recent models, such as LLaMA-VID~\cite{li2023llama-vid} and Video-LLaMA~\cite{zhang2023video-llama} propose strategies for linking visual features to language prompts, harnessing the capabilities of LLaMA~\cite{touvron2023llama, touvron2023llama2} models for video description. 
Furthermore, VideoChat~\cite{li2023videochat, li2023videochat2} treats video captioning as a subtask of visual question answering, while StreamingCaption~\cite{zhou2024streaming} can generate captions for streaming videos using a memory mechanism.

Notably, the AutoAD series~\cite{han2023autoad, han2023autoad2, han2024autoad3} apply video captioning to a specific domain -- synthesizing descriptive narrations for movie scenes to assist visually impaired individuals in watching movies.
Similarly, in the context of soccer, a distinctive sports scenario, we develop a tailored soccer game commentary model to enrich the viewing experience for audiences.

\vspace{2pt}
% \noindent{\textbf{Sports Video Understanding.}}
\noindent{\textbf{Sports video understanding}}~\cite{thomas2017computervisionsports} has widely attracted the interest of researchers due to its complexity and professional relevance.
Early works such as FineGym~\cite{shao2020finegym} and FineDiving~\cite{xu2022finediving} aim to develop fine-grained datasets for action recognition and understanding in specific sports.
Subsequently, focusing on soccer, a series of SoccerNet~\cite{giancola2018soccernet1} datasets systematically address various challenges related to soccer, including player detection~\cite{vandeghen2022semi}, action spotting~\cite{giancola2018soccernet1}, replay grounding~\cite{held2023vars}, player tracking~\cite{cioppa2022soccernet-track}, camera calibration~\cite{giancola2018soccernet2} and re-identification~\cite{deliege2021soccernet3}.
These endeavours have paved the way for more ambitious research goals, such as utilizing AI for soccer game commentary~\cite{densecap, qi2023goal}.
Additionally, other approaches have targeted aspects of sports analysis, such as basketball game narration~\cite{yu2018fine} and tactics analysis~\cite{wang2024tacticai}.

A concurrent work, SoccerNet-Echoes~\cite{gautam2024soccernet-echoes} proposes to leverage audio from videos for ASR and translation to obtain richer text commentary data. However, this approach overlooks that unprocessed audios often contain non-game-related utterances, which may confuse model training.
Building upon the aforementioned progress, our goal is to construct a dataset with improved alignment to train a more professional soccer game commentary model, thereby achieving a better understanding of sports video.

\section{Conclusion}

In this paper, we consider a highly practical and commercially valuable task: automatically generating professional textual commentary for soccer games.
% to offer audiences a better viewing experience.
Specifically, we have observed a prevalent misalignment between visual contents and textual commentaries in existing datasets.
To address this, we manually correct the timestamps of textual commentary in 49 videos in the existing dataset, establishing a new benchmark for the community, termed as {\bf SN-Caption-test-align}. 
Building upon the manually checked data, we propose a multi-modal temporal video-text alignment pipeline that automatically corrects and filters existing data at scale, 
which enables us to construct a higher-quality soccer game commentary dataset, named {\bf MatchTime}. Based on the curated dataset, we present \textbf{MatchVoice}, a soccer game commentary model, which can accurately generate professional commentary for given match videos, significantly outperforming previous methods. Extensive experiments have validated the critical performance improvements achieved through data alignment, as well as the superiority of our proposed alignment pipeline and commentary model.

% In this paper, we consider a highly practical and commercially valuable task: synthesizing professional commentary for soccer games to provide the audience with a better viewing experience. 
% Specifically, we have observed significant visual-text misalignment in existing datasets and manually annotated 49 matches to create a stronger commentary benchmark. 
% Built upon this, we design a multimodal temporal video-text alignment pipeline named {\em \textbf{MatchTime}} to obtain a higher-quality dataset. 
% Leveraging the power of LLMs, we develop a LLaMA-based multimodal commentary model for soccer, named {\em \textbf{MatchVoice}}.
% Extensive experiments demonstrate that our model, trained on the aligned dataset, exhibits superior performance on our manually corrected benchmark, indicating the essentiality of better data alignment.
\section*{Limitations}
Although our proposed {\bf MatchVoice} model can generate professional textual commentary for given soccer game videos, it still inherits some limitations from existing data and models: 
(i) Following previous work, our commentary remains anonymous and cannot accurately describe player information on the field. 
This is left for future work, where we aim to further improve the dataset and incorporate knowledge and game background information as additional context; 
and 
(ii) {\bf MatchVoice} may sometimes struggle to distinguish between highly similar actions, such as {\em corner kicks} and {\em free kicks}. 
This mainly stems from the current frozen pre-trained visual encoders and language decoders. 
Our preliminary findings suggest that fine-tuning on soccer-specific data might effectively address this issue in the future.

\section*{Acknowledgments}
This work is funded by National Key R\&D Program of China (No.2022ZD0161400).

%\bibliography{anthology, custom}
\bibliography{main}

\clearpage
\appendix

\section{Appendix}
\label{sec:appendix}

% \weidi{there are quite a lot of details missing here, 
% for example, the prompt used for sentence similarity,
% the prompt for GPT-score, etc.}

% \weidi{I'll do another pass after you fix all the comments.}

\subsection{Dataset Split}

We split the total 471 matches of our dataset (including automatically aligned {\bf MatchTime} and manually curated {\bf SN-Caption-test-align benchmark}) into training (373 matches), validation (49 matches), and test (49 matches) sets, consisting of 26,058, 3,418, and 3,267 video clip-text pairs, respectively. 
Notably, all test samples are from our manually checked {\bf SN-Caption-test-align}, which serves as a better benchmark on soccer game commentary generation for the community.

\subsection{Implementation Details}
In this section, we provide additional details regarding the implementations as follows.

\vspace{3pt} \noindent{\textbf{Baseline Methods.}}
For baselines, we retrain several variants of SN-Caption~\cite{densecap} with its official implementation.
NetVLAD++~\cite{netvlad} is adopted to aggregate the temporal information of the extracted features.
Then the pooled features are decoded by an LSTM~\cite{hochreiter1997long}.

\vspace{3pt} \noindent{\textbf{Event Summarization.}}
\label{appendix:event_summarization}
Considering that the narrations by commentators may be fragmented and colloquial, we feed the ASR-generated narration texts into the LLaMA-3~\cite{llama3} model and use the following prompt to summarize them into event descriptions for every 10 seconds:

\begin{mdframed}[backgroundcolor=gray!10] 
{\em 
"I will give you an automatically recognized speech with timestamps from a soccer game video. The narrator in the video is commenting on the soccer game. Your task is to summarize the key events for every 10 seconds, each commentary should be clear about the person name and soccer terminology. Here is this automatically recognized speech: \textbackslash n \textbackslash n \{timestamp intervals: ASR sentences\} \textbackslash n \textbackslash n You need to summarize 6 sentence commentaries for 0-10s, 10-20s, 20-30s, 30-40s, 40-50s, 50-60s according to the timestamps in automatically recognized speech results, every single sentence commentary should be clear and consise about the incidents happened within that 10 seconds for around 20-30 words. Now please write these 6 commentaries.\textbackslash n Answer:"}
\end{mdframed}

\vspace{2pt}
\noindent{\textbf{Timestamp Prediction.}}
\label{appendix:timestamp_prediction}
With the event descriptions and their corresponding timestamps, we input them along with the textual commentaries into LLaMA-3~\cite{llama3} to predict the timestamps for the textual commentaries based on sentence similarity, providing a solid foundation for fine-grained alignment.
The prompt used for this step is as follows:

\begin{mdframed}[backgroundcolor=gray!10] 
{\em 
"I have a text commentary of a soccer game event at the original time stamp: \textbackslash n \textbackslash n{Original timestamp here}: \{Original commentary here (from SoccerNet-Caption)\} \textbackslash n \textbackslash n and I want to locate the time of this commentary among the following events with timestamp: \textbackslash n \{timestamp intervals of 10s: summarized events\}. \textbackslash n These are the words said by narrator and I want you to temporally align the first text commentary according to these words by narrators since there is a fair chance that the original timestamp is somehow inaccurate in time. So please return me with a number of time stamp that event is most likely to happen. I hope that you can choose a number of time stamp from the ranges of candidates. But if really none of the candidates is suitable, you can just return me with the original time stamp. Your answer is:"
}
\end{mdframed}

\subsection{Evaluation Metrics}
\label{appendix:evaluation_metrics}
In this paper, most evaluation metrics (BLEU~\cite{papineni2002bleu}, METEOR~\cite{banerjee2005meteor}, ROUGE-L~\cite{lin2004rouge}, CIDEr~\cite{vedantam2015cider}) are calculated using the same function settings with SoccerNet-Caption~\cite{densecap}, by the implementation of \textit{pycocoevalcap} library.
GPT-score~\cite{fu2023gptscore} is given by GPT-3.5 with the following text as prompt: 

\begin{mdframed}[backgroundcolor=gray!10] 
{\em
"You are a grader of soccer game commentaries.
There is a predicted commentary by AI model about a soccer game video clip and you need to score it comparing with ground truth. 
\textbackslash n \textbackslash n 
You should rate an integer score from 0 to 10 about the degree of similarity with ground truth commentary (The higher the score, the more correct the candidate is). 
You must first consider the accuracy of the soccer events, then to consider about the semantic information in expressions and the professional soccer terminologies. 
The names of players and teams are masked by "[PLAYER]" and "[TEAM]". 
\textbackslash n \textbackslash n 
The ground truth commentary of this soccer game video clip is: 
\textbackslash n \textbackslash n
"\{Ground truth here.\}"
\textbackslash n \textbackslash n
I need you to rate the following predicted commentary from 0 to 10:
\textbackslash n \textbackslash n
"\{Predicted Commentary here.\}"
\textbackslash n \textbackslash n
The score you give is (Just return one number, no other word or sentences):"}
\end{mdframed}

\subsection{Details of Temporal Alignment}
\label{appendix:temporal_alignment}
For our proposed fine-grained temporal alignment model, sampling appropriate positive and negative examples for contrastive learning affects the results. 

\begin{table}[h]
    \centering
    % \footnotesize
    \small
    \tabcolsep=0.15cm
     \begin{tabular}{c|ccccc}
     \toprule
    \multicolumn{1}{c|}{Window(s)} & 60 & 120 & 150 & 180 & 240 \\
          \midrule
  $\mathrm{avg} (\Delta)$ (s)        & -0.54 & \textbf{0.03} & 0.44 & 2.34 & -5.77 \\
  $\mathrm{avg} (| \Delta |)$ (s)    & 14.06 & \textbf{6.89} & 15.06 & 11.94 & 16.77 \\
  \midrule
  $\mathrm{window}_{10}$ (\%)        & 97.71 & \textbf{98.17} & 91.28 & 91.28 & 85.78 \\
  $\mathrm{window}_{30}$ (\%)        & 94.04 & \textbf{95.41} & 88.07 & 88.53 & 82.57 \\ 
  $\mathrm{window}_{45}$ (\%)        & 81.65 & \textbf{91.28} & 84.40 & 83.94 & 81.65 \\ 
  $\mathrm{window}_{60}$ (\%)        & 59.17 & \textbf{80.73} & 75.23 & 79.36 & 78.90 \\ 
% Mean: -5.76605504587156 Abs Mean: 16.76605504587156
% In 60: 85.78, In 45: 82.57, In 30: 81.65, In 10: 78.90
  \bottomrule
        \end{tabular}
        \vspace{-4pt}
        \caption{\textbf{Alignment Results of Different Windows}}
        \vspace{-4pt}
        \label{tab:contrastive_window}
\end{table}

As depicted in Table \ref{tab:contrastive_window}, we have experimented with sampling windows of different lengths and observed that using a 120-second window around the manually annotated ground truth ({\em i.e.}, 60 seconds before to 60 seconds after) can yield optimal alignment performance. 
Specifically, for each text commentary, 
we treat the key frame corresponding to its ground truth timestamp as the positive sample, 
while other samples within a fixed window size, sampled at 1 FPS, serve as negative samples~({\em i.e.}, those within 5 to 60 seconds temporal distance to the ground truth timestamp).

Considering that data pre-processing based on ASR and LLM provides a coarse alignment and that there might be replays in soccer game videos, during the inference stage, we use key frames from 45 seconds before to 30 seconds after the current textual commentary timestamp as candidates.

% Additional experiments have been conducted to identify the optimal window size for sampling features in our designed fine-grained temporal alignment based on contrastive learning 
% The results, presented in Table \ref{tab:contrastive_window}, indicate that the most effective alignment is achieved when visual features are sampled within a 120-second window surrounding the timestamp (60s before and 60s after).

\subsection{Divergence Among Annotators}
Although the recruited volunteers are all football enthusiasts, there exists noticeable subjectivity and variability in manual annotations due to different understandings of soccer terminology and actions. 

To quantify this, three volunteers are asked to annotate two matches from our {\bf SN-Caption-test-align} benchmark. 
We observe an ``alignable/unalignable'' disagreement among different annotators on \textbf{6.29\%} of the samples. Additionally, the average of maximum discrepancy between the timestamps provided by different annotators is \textbf{5.57} seconds, which can somehow seen as the performance upper-bound of automatic alignment models.

% Even though the enlisted volunteers are all aficionados familiar with soccer, the subjective nature of manual annotation is still evident, which is particularly pronounced due to discrepancies in the understanding of soccer terminology and actions among different individuals. 
% In an effort to quantify this variability, two matches from the \textit{SN-Caption-test-align} dataset were selected for annotation by a cohort of diverse volunteers. 
% It reveals that there was a divergence between \textit{alignable/unalignable} of \textbf{6.29\%} of the commentary. 
% Furthermore, there is a maximum temporal discrepancy averaging \textbf{5.57} seconds among the annotators. Such variance between different annotators could be regarded as the ceiling level of our alignment model.

\subsection{More Qualitative Results}
% In this part, we provide some qualitative results of our alignment model in Figure~\ref{fig:qualitative_alignment_appendix}. 
% Moreover, we present more qualitative results of our proposed \textbf{MatchVoice} model on soccer game commentary generation, shown in Figure~\ref{fig:qualitative_commentary_appendix}.

\begin{figure*}[th]
    \centering
    \includegraphics[width=\textwidth]{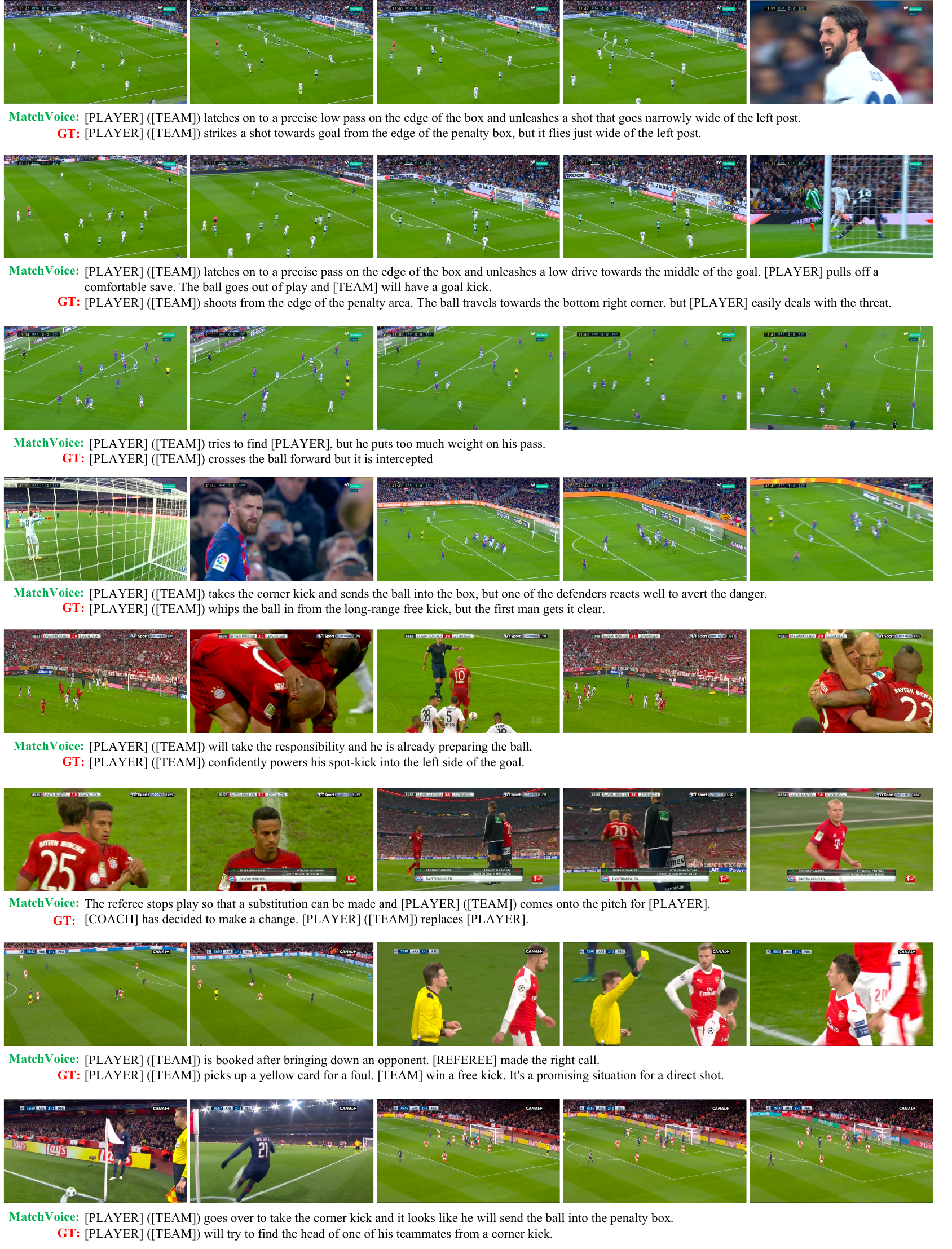}
    \caption{\textbf{More qualitative results on commentary generation.}}
    \label{fig:qualitative_commentary_appendix}
    \end{figure*}

In this part, we present more qualitative results of our proposed \textbf{MatchVoice} model on soccer game commentary generation, shown in Figure~\ref{fig:qualitative_commentary_appendix}.

% \begin{figure}[b!]
%     \centering
%     \includegraphics[width=0.5\textwidth]{latex/images/appendix_align.pdf}
%     \caption{\textbf{More qualitative results on Temporal Alignment.} Timestamps before alignment are in \textcolor{orange}{Orange}, those after alignment are in \textcolor[rgb]{0,0.5,0}{Green}.}
%     \label{fig:qualitative_alignment_appendix}
%     \end{figure} 
    
\end{document}